\documentclass[conference]{IEEEtran}
\usepackage{cite}

\ifCLASSINFOpdf
  \usepackage[pdftex]{graphicx}
\else
  \usepackage[dvips]{graphicx}
\fi

\usepackage{placeins}
\usepackage{booktabs}
\usepackage{siunitx}

\usepackage[cmex10]{amsmath}

\usepackage{algorithm}
\usepackage[noend]{algpseudocode}

\usepackage{array}

\usepackage{booktabs} 
\usepackage{enumitem}
\usepackage{algorithm}
\usepackage[noend]{algpseudocode}
\usepackage{graphicx,calc}

\usepackage[numbers,sort]{natbib}

\usepackage{url}

\usepackage{color}
\newlength\myheight
\newlength\mydepth
\settototalheight\myheight{Xygp}
\usepackage{multirow}
\usepackage{pbox}

\usepackage{url}

\usepackage[font=footnotesize]{caption}

\usepackage[colorinlistoftodos]{todonotes}

\newcommand{\ignore}[1]{}

\begin{document}
\title{Automated Curriculum Learning by Rewarding Temporally Rare Events}


\author{
\IEEEauthorblockN{Niels Justesen}
\IEEEauthorblockA{IT University of Copenhagen\\
Copenhagen, Denmark\\
noju@itu.dk}
\and
\IEEEauthorblockN{Sebastian Risi}
\IEEEauthorblockA{
IT University of Copenhagen\\
Copenhagen, Denmark\\
sebr@itu.dk}

}

\maketitle

\begin{abstract}
Reward shaping allows reinforcement learning (RL) agents to accelerate learning by receiving additional reward signals. However, these signals can be difficult to design manually, especially for complex RL tasks. We propose a simple and general approach that determines the reward of pre-defined events by their rarity alone. Here events become less rewarding as they are experienced more often, which encourages the agent to continually explore new types of events as it learns. The adaptiveness of this reward function results in a form of automated curriculum learning that does not have to be specified by the experimenter. We demonstrate that this \emph{Rarity of Events} (RoE) approach enables the agent to succeed in challenging VizDoom scenarios without access to the extrinsic reward from the environment. Furthermore, the results demonstrate that RoE learns a more versatile policy that adapts well to critical changes in the environment. Rewarding events based on their rarity could help in many unsolved RL environments that are characterized by sparse extrinsic rewards but a plethora of known event types. 
\end{abstract}

\section{Introduction}
Deep reinforcement learning and deep neuroevolution have achieved impressive results learning to play video games \cite{justesen2017deep} and controlling both simulated and physical robots \cite{chebotar2017combining, mirowski2016learning, andrychowicz2017hindsight, gu2016continuous}. These approaches, however, struggle to learn in environments where feedback signals (also called rewards) are sparse and/or delayed. A popular way to overcome this issue is to shape the reward function with prior knowledge such that the agent receives additional rewards to guide its learning process \cite{ng2003shaping, laud2004theory, lample2017playing}. Another approach is to gradually increase the difficulty of the environment to ease learning through curriculum learning \cite{bengio2009curriculum, wu2016training}. Both approaches are time-consuming, require substantial domain knowledge and are especially difficult to implement for complex environments. In this paper, we propose a simple method that automatically shapes the reward function during training and performs a form of curriculum learning that adapts to the agent's current performance. The only required domain knowledge is the specification of a set of positive events that can happen in the environment (e.g.\ picking up items, moving, winning etc.), which is easy to implement if raw state changes are accessible. 

The method introduced in this paper rewards a reinforcement learning (RL) agent by the rarity of experienced events such that rare events have a higher value than frequent events. The idea is to completely discard the extrinsic reward and instead motivate the agent intrinsically toward a behavior that explores the pre-defined events. As the agent first experiences certain types of events that are relatively easy to learn (e.g.\ moving around and picking up items) they will slowly become less rewarding, pushing the agent to explore rare and potentially more difficult events. Thus by only rewarding events for their rarity, the system performs a form of automated curriculum learning.  

The goal of this approach is to learn through a process of  \emph{curiosity} rather than optimizing toward a difficult pre-defined goal. We apply our method, called \emph{Rarity of Events} (RoE), to learn agent behaviors from raw pixels in the VizDoom framework \cite{kempka2016vizdoom}. While our approach could be applied to any reward-based learning method and possibly also fitness-based evolutionary methods, in this paper we train deep convolutional networks through the actor-critic algorithm A2C \cite{mnih2016asynchronous}. In the future, RoE could offer a new way to learn versatile behaviors in increasingly complex environments such as StarCraft, which is a yet unsolved reinforcement learning problem \cite{vinyals2017starcraft}.

The paper is structured as follows. We first review relevant previous work, including related approaches in Section~\ref{sec:previous}. After explaining RoE (Section~\ref{sec:approach}), we demonstrate the usefulness of the method on five challenging VizDoom scenarios with sparse rewards and show how RoE learns a versatile behavior that can adapt to critical changes in the environment (Section~\ref{sec:results}). 

\section{Previous Work}
\label{sec:previous}

\subsection{Deep Reinforcement Learning}
Deep reinforcement learning allows learning agent behaviors in video games directly from screen pixels, including Atari games \cite{mnih2015human}, first-person shooters \cite{lample2017playing, wu2016training,kempka2016vizdoom}, and car racing games \cite{mnih2016asynchronous}. These methods are typically variants of Deep Q Networks (DQN) \cite{mnih2015human} or actor-critic methods with  parallel actor-learners such as Asynchronous Advantage Actor-Critic (A3C) \cite{mnih2016asynchronous}. Neuroevolution \cite{floreano2008neuroevolution,risi2017neuroevolution} has also recently shown promising results in playing Atari games and can be easier to parallelize  \cite{salimans2017evolution, such2017deep}. 

A key requirement for deep RL methods to work out of the box are frequent and easy obtainable reward signals from the environment that can guide learning toward an optimal behavior. An infamous Atari game where this is not the case is Montezuma's Revenge; for this game with very sparse rewards, both DQN and A3C variants fail \cite{mnih2015human, mnih2016asynchronous}. 

The lack of frequent reward signals can be overcome by reward shaping, where a smoother reward function  is designed using prior domain knowledge \cite{laud2004theory, lample2017playing}, or by gradually increasing the difficulty of the environment (e.g.\ the level itself or the NPCs' behaviors) to ease learning through curriculum learning \cite{bengio2009curriculum, wu2016training}. Related to curriculum learning is a method called \emph{Power Play} that searches for new unsolvable problems while the agent is trained to progressively match the difficulty of the environment \cite{schmidhuber2013powerplay}. Another related approach is hierarchical reinforcement learning where a meta-controller controls one or more sub-policies that are trained to reach sub-goals (equivalent to events) \cite{kulkarni2016hierarchical, botvinick2009hierarchically}.

\subsection{Curiosity \& Intrinsic Motivation}
In curiosity-driven learning the agent seeks to explore new situations guided by intrinsic motivation \cite{oudeyer2018computational, ryan2000intrinsic, kaplan2007intrinsically}. One theory of intrinsic motivation is \emph{reduction of cognitive dissonance}, i.e.\ the motivation to learn a cognitive model that can explain and predict sensory input \cite{oudeyer2009intrinsic, festinger2017theory}. This theory has also been formalized in the context of RL in which agents are intrinsically rewarded when observing temporarily novel, interesting, or surprising patterns based on their own world model \cite{schmidhuber2010formal}. A related idea is \emph{optimal incongruity}, where discrepancies between the currently perceived and what is usually perceived produce a high stimulus; thus novel situations that yet lie within our current understanding are highly rewarding \cite{hunt1965intrinsic, berlyne1960conflict}. The prediction error of a learning model can thus be used directly to define the reward function \cite{gordon2012hierarchical}.

One way of implementing intrinsic motivation is to model the expected learning progress $\zeta(s,a)$ of a state-action pair \cite{lopes2012exploration}. The Intrinsic Curiosity Module (ICM) is another approach that encodes states $s_{t}$ and $s_{t+1}$ into features $\Phi(s_{t})$ and $\Phi(s_{t+1})$ and determines the intrinsic reward based on the prediction error of these features and the forward model's features \cite{pathak2017curiosity}. State-density models that assign probabilities to screen images, can be learned together with a policy and then determine intrinsic motivation as the model's temporal change in prediction, such that surprising screen images produce higher rewards \cite{bellemare2016unifying}. 

Rewarding RL agents based on the novelty of events has been explored earlier with tabular Q-learning in a simple 3D environment \cite{maher2008achieving}, where the reward is highest when novelty is moderate. A combination of habituation theory and self-organizing maps was employed to vary the agent's curiosity (the reward signal toward certain events).

\subsection{Novelty Search}
The pursuit of novel situations also shares some similarities with novelty search \cite{lehman2011abandoning} in evolutionary computation. The idea of novelty search is to search for novel behaviors instead of optimizing toward a specific objective directly. Both novelty search and our approach RoE push the search toward unexplored areas; however, novelty search does so for a population of individuals where novelty is defined as the behavioral distance to other behaviors in the population. Our approach is trained through reinforcement learning and novelty (or rather rarity of events) is based on experiences of previous versions of the policy. 


\subsection{VizDoom}
The approach in this paper is tested in VizDoom, an AI research platform based on the commercial video game Doom that allows learning from raw visual information \cite{kempka2016vizdoom}. The VizDoom framework includes several diverse environments, some of which are very challenging to learn due to their sparse and delayed rewards. 
Several deep RL approaches have been applied to Doom, which 
include auxiliary learning \cite{kulkarni2016deep, lample2017playing}, game-feature augmentation \cite{bhatti2016playing, chaplot2017arnold, dosovitskiy2016learning}, manual reward shaping \cite{chaplot2017arnold, lample2017playing, dosovitskiy2016learning}, and curriculum learning \cite{wu2016training}. A very different approach by Alvernaz and Togelius applies neuroevolution on top of a pre-trained auto-encoder \cite{alvernaz2017autoencoder}. 
In this paper, we purposefully build on a vanilla implementation of the RL  algorithm A2C, to set a baseline for how well RoE can help in challenging VizDoom scenarios. 

\section{Approach}
\label{sec:approach}
This section describes our \emph{Rarity of Events} (RoE) approach and its integration with A2C in VizDoom.
\subsection{Rewarding Temporally Rare Events}
\label{sec:rare-events}
The reward function in RoE adapts throughout training to the policy's ability to explore the environment. By rewarding events based on how often they occur during training, the agent is intrinsically motivated toward exploring new parts of the environment rather than aiming for a single goal that might be difficult to obtain directly. In effect, the approach performs a form of curriculum learning since events are rewarded based on the agent's current ability to obtain them. As the agent learns, it becomes less interested in events that are frequent and \emph{curious} about newly discovered events.

Our method requires a set of pre-defined events, and the reward $R_{t}(\epsilon_{i})$ for experiencing one of these events $\epsilon_{i}$ at time $t$ is determined by its temporal rarity $\frac{1}{\mu_{t} (\epsilon_{i})}$, where $\mu_{t} (\epsilon_{i})$ is the temporal episodic mean occurrence of $\epsilon_{i}$ at time $t$, i.e. how often $\epsilon_{i}$ occurs per episode at the moment. The mean occurrences of events are clipped to be above a lower threshold $\tau$ (we used 0.01 such that the maximum reward for any event is 100). For a vector of event occurrences $x$, such that $x_{i}$ is the number of times $\epsilon_{i}$ occurred in a game step, the reward is the sum of all event rewards:
\begin{equation}
R_{t}(x) = \sum_{i=1}^{\vert x \vert} x_{i} \frac{1}{\textrm{max}(\mu_{t} (\epsilon_{i}), \tau)}.
\label{eq:rarity}
\end{equation}

The rarity measure $\frac{1}{\mu_{t} (\epsilon_{i})}$ is not arbitrary but is designed such that all events have equal importance. If any event $\epsilon_{i}$ is experienced $n$ times during an episode, and $n=\mu_{t} (\epsilon_{i})$ (which is the expected amount), then the accumulated reward for $\epsilon_{i}$ is 1 regardless of the rarity. This means that in theory all events have equal importance. In practice, the policy might learn that some events have a negative or positive influence on the occurrence of others.

\subsection{Determining the Temporal Episodic Mean Occurrence}
There are arguably many ways to determine the temporal episodic mean occurrence $\mu_{t} (\epsilon_{i})$; here we employ a simple approach that nevertheless achieves the desired outcome. Whenever an episode during training reaches a terminal state, a vector $\epsilon$ containing the occurrence of events in this episode is added to a buffer of size $N$. The size of the buffer determines the adaptability of the reward function. If $N$ is small, the agent quickly becomes \emph{bored} of new events as it easily forgets their rarity in the past. If $N$ is large, the agent will stay \emph{curious} for a longer period of time. The temporal episodic mean occurrence $\mu_{t} (\epsilon)$ is then determined as the mean of all records in the buffer, i.e.\ the episodic mean of the last $N$ episodes.  

\subsection{Events in Doom}
We track 26 event types in VizDoom by implementing a function that determines  which events occur in every state transition (i.e.\ in each time step). The event types include movement (one unit), shooting (decrease in ammo), picking up an item (one event for each item type; health pack, armor, ammo, and weapons 0--9), killing (one for each weapon type 0--9 as well as one regardless of weapon type). Movement events are triggered when the agent has traveled one unit from the position of the last movement event (or the initial position if the agent has not yet moved). 

\subsection{Policy}
The presented reward shaping approach can be applied to most (if not all) RL methods that learn from a reward signal. It could potentially also be applied to evolutionary approaches such as Evolution Strategies by defining fitness as the sum of rewards in an episode. 
A standard policy network is employed that has  three convolutional layers followed by a fully connected layer of 512 units, and a policy and value output. We use filter sizes of [32, 64, 32] with strides [4, 2, 1], ReLU activations for hidden layers, and softmax for the policy output. 

The input is a single frame of 160$\times$120 pixels in grayscale, cropped by removing 10 pixels on top/bottom and 30 pixels on the sides and then resized to 80$\times$80. 
In most of the scenarios, the agent can perform four actions: attack, move forward, turn left, and turn right. In this case, the policy output has $2^{4}=16$ values to allow any combination of the four actions. The event buffer is updated whenever a worker reaches a terminal state. The rewards from VizDoom, which vary between -100 and 100, are normalized to $[0, 1]$. Rewards based on our approach are not normalized and vary between $[0, 100]$ (due to $\tau=0.01$), while for all events where $\mu_{t} (\epsilon_{i}) \geq 1$ the reward will be between 0 and 1 (following Equation~\ref{eq:rarity} in  Section~\ref{sec:rare-events}).

\subsection{Advantage Actor-Critic (A2C)}
\label{sec:a2c}
The deep networks in this paper are trained with the deep reinforcement learning algorithm A2C, a synchronous variant of Asynchronous Advantage Actor-Critic (A3C) \cite{mnih2016asynchronous}, which is able to reach state-of-the-art performance in a wide range of environments \cite{wu2017scalable, wang2016learning, schulman2017proximal}. 

A2C is an actor-critic method that optimizes both a policy $\pi$ (the actor) and an estimation of the state-value function $V(s)$ (the critic). Parallel worker threads share the same model parameters and synchronously collect trajectories $(s_{t}, s_{t+1}, a_t, r_{t+1})$ for $t_{max}$ game steps where after the model's parameters are updated. Threads restart new episodes individually when they are done. The discounted return $R_t = \sum_{i=1}^{k-1} \gamma^{i} r_{t+i} + \gamma^{k} V(s_{t+k})$, where $k$ is the number of trajectories collected after $t$, and the advantage $A(s_t, a_t)=R_t - V(s_t)$ is determined for each step, for every worker. A2C then uses the traditional A3C update rules in \cite{mnih2016asynchronous} based on the policy loss $\log\pi(a_i|s_i) A(s_i)$ and value loss; the mean squared error between the experienced $R_t$ and the predicted $V(s_t)$: $\frac{1}{2} (R_t - V(s_t))^{2}$. In contrast to A3C, A2C updates the parameters synchronously in batches.


\begin{center}
\begin{figure}[!htb]
  \hspace*{-0.12in}
  \includegraphics[width=1.02\columnwidth]{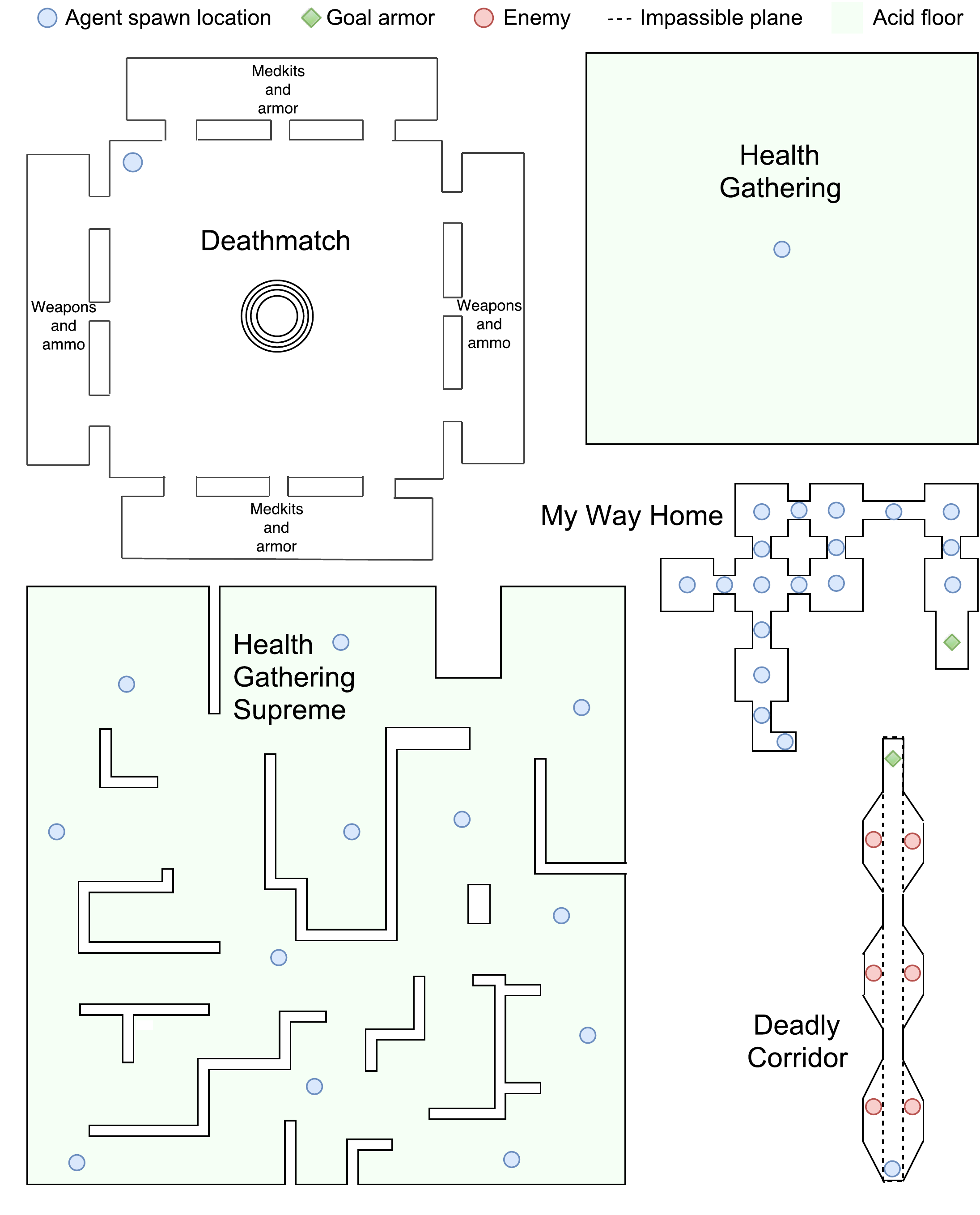} 
  \caption{The five ViZdoom scenarios. Scenarios with multiple spawning positions randomly select one of them at the start of an episode. The episode ends when the goal armor, which only appears in \emph{My Way Home} and \emph{Deadly Corridor}, is picked up. The agent periodically looses health when standing on acid floors.} 
  \label{fig:scenarios}
  \vspace*{-0.0in}
\end{figure}
\end{center}

\section{VizDoom Testing Scenarios}
\label{sec:scenarios}
This section describes the five VizDoom scenarios used in our experiments. They all have sparse and/or delayed rewards and are therefore a good test domain for our approach. The scenarios are from the original VizDoom \cite{kempka2016vizdoom} repository\footnote{https://github.com/mwydmuch/ViZDoom/tree/master/scenarios}. 

For each scenario we also detail the extrinsic reward from the environment, which is used when training models without RoE. 
Some of these extrinsic rewards were rescaled to be coherent across scenarios. If not stated otherwise, the agent can move forward, turn left, turn right, and shoot. Screenshots from these scenarios are shown in Figure~\ref{fig:screenshots}, with top-down views in Figure~\ref{fig:scenarios}. 

\subsubsection{Health Gathering}
The goal is to survive as long as possible in a square room with an acid floor that deals damage periodically. Medkits spawn randomly in the room and can help the agent to survive as they heal when picked up. The agent is rewarded 1 for every time step it is alive, and -100 for dying. The maximum episode length is 2,100 time steps. The agent cannot shoot.

\subsubsection{Health Gathering Supreme}
Same as \emph{Health Gathering} but within a maze.

\subsubsection{My Way Home}
The goal is to pick up an armor, which gives a reward of 100 and ends the scenario immediately. The agent cannot shoot and is rewarded -0.1 for every time step it is alive. The agent starts an episode at one of the randomly chosen spawn locations with a random rotation.

\subsubsection{Deadly Corridor}
Similarly to \emph{My Way Home}, the goal is to pick up an armor, which gives a reward of 100 and ends the scenario immediately. The armor is located at the end of a corridor, which is guarded by enemies on both sides. The agent must kill most, if not all of the enemies to reach it, and receives a -100 reward if it dies. The original reward shaping function (the distance to the armor) has been removed to make it harder and to compare RoE with a baselines that does not use any reward shaping. The maximum episode length is 2,100 time steps. 

\begin{center}
\begin{figure}[!htb]
  \includegraphics[width=1\columnwidth]{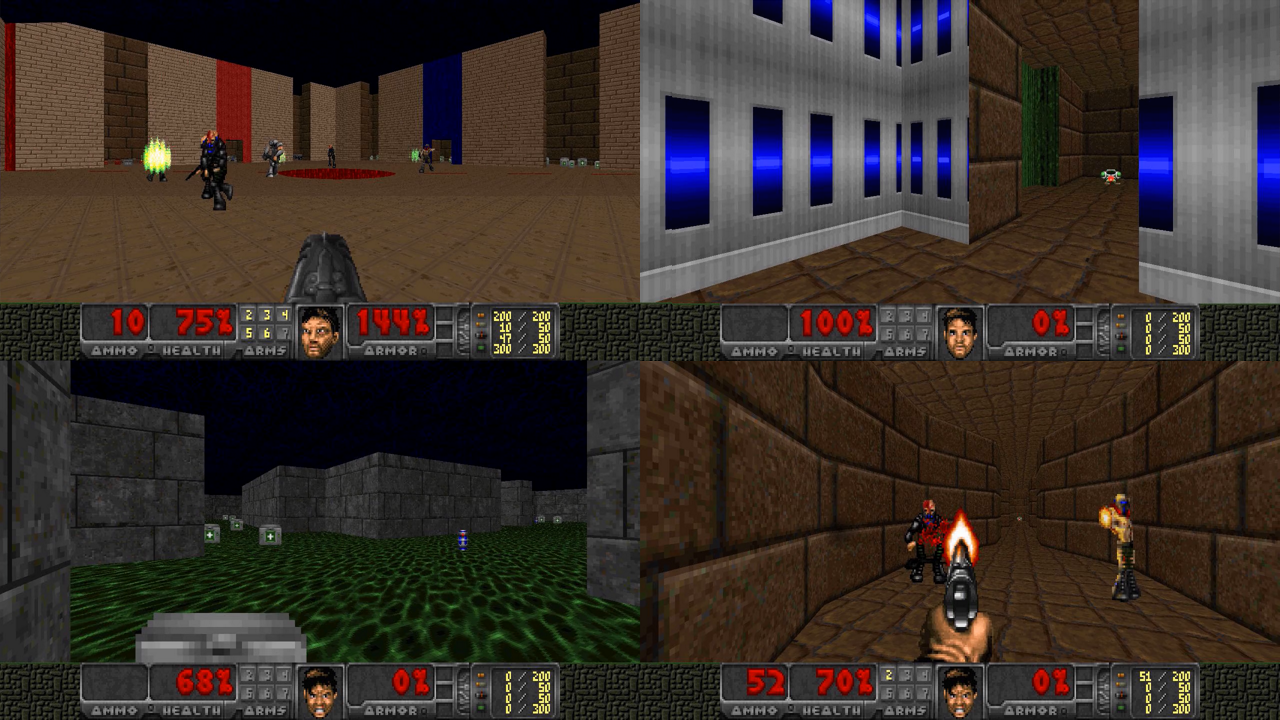} 
  \caption{From top-left to bottom-right: Screenshot from \emph{Deathmatch}, \emph{My Way Home}, \emph{Health Gathering Supreme}, and \emph{Deadly Corridor}. Notice that in some scenarios the agent cannot shoot. The scenario \emph{Health Gathering} is similar to \emph{Health Gathering Supreme} but without walls within the room.} 
  \label{fig:screenshots}
\end{figure}
\vspace*{-0.3in}
\end{center}

\subsubsection{Deathmatch}
The agent spawns in a large battle arena with an open area in the middle and four rooms, one in each direction that contain either medkits and armor, or weapons (chainsaw, super shotgun, chaingun, rocket launcher, and plasma gun) and ammunition for each weapon. The maximum episode length is 4,200 time steps. The agent is rewarded the following amounts  when killing an enemy: Zombieman (100), ShotgunGuy (300), MarineChainsawVzd (300), Demon (300), ChaingunGuy (400), HellKnight (1,000). These enemies spawn randomly on the map when the scenario  starts. 

To test how well the approach can adapt to new scenarios, five variations of \emph{Deathmatch} were also created that only include a certain weapon type. 
These scenarios are called \emph{Deathmatch Chainsaw}, \emph{Deathmatch Chaingun}, \emph{Deathmatch Shotgun}, \emph{Deathmatch Plasma}, and \emph{Deathmatch Rocket} to denote which weapon that remains on the map. The ammunition for the other weapons was also removed.

\section{Results}
\label{sec:results}
We tested A2C with our approach \emph{Rarity of Events} (A2C+RoE) on the five  VizDoom scenarios described in Section~\ref{sec:scenarios}. The \emph{Deathmatch} variations were not used for training. As a comparison baseline, A2C was also trained using the extrinsic reward from the environment as described in Section \ref{sec:scenarios}. Due to computational constraints we only trained each method once on each scenario.

When training with A2C+RoE, the agent did not have access to the extrinsic reward throughout training but only the intrinsic reward based on the temporal rarity of the pre-defined events. The algorithms ran for $10^{7}$ time steps for each scenario and $7.5 \times 10^{7}$ for the Deathmatch scenario. For both A2C and A2C+RoE we save a copy of the model parameters whenever the mean extrinsic reward across all workers improves. The last copy is considered to be the final model that we use in our tests. The complete configurations for A2C and A2C+RoE are shown in Table~\ref{tab:config} and the code for the experiments and trained models are available on GitHub\footnote{\url{https://github.com/njustesen/rarity-of-events}}. Videos of the learned policies are available on YouTube\footnote{\url{https://youtu.be/YG-lf732a0U}}.

\begin{table}[t]
  \centering
  \begin{tabular}{l*{2}{c}r}
    \hline
    \multicolumn{2}{c}{A2C} \\
    \hline
    \hline
    Learning rate & 7e-4 \\
    $\gamma$ (discount factor) & 0.99 \\
    Entropy coefficient & 0.01 \\
    Value loss coefficient & 0.5 \\
    Learning rate & 0.0007 \\
    Max. gradient-norm & 0.5\\
    Worker threads & 4 (16 in DM)\\
    $t_{max}$ (Steps per. update) & 20\\
    Batch size & 64\\
    Frame skip & 4\\
    \hline
    \multicolumn{2}{c}{RMSprop Optimizer} \\
    \hline
    \hline
    $\epsilon$ & 1e-5 \\
    $\alpha$ & 0.99 \\
    \hline
    \multicolumn{2}{c}{RoE} \\
    \hline
    \hline
    $N$ (event buffer size) & 100 \\
    $\tau$  (mean threshold) & 0.01 \\
    \hline
    \end{tabular}
\caption{Experimental configurations for A2C and A2C+RoE. 16 worker threads were used in \emph{Deathmatch}. }
\label{tab:config}
\end{table}

\begin{table}[t]
  \centering
  \vspace*{0.2in}
  \begin{tabular}{l*{2}{c}|c}
    Scenario & A2C & A2C+RoE & t-test\\
    \hline
    \hline
    Health Gathering & 399 $\pm$ 107 & \textbf{1261} $\pm$ 533 & $p<0.0001$ \\
    Health Gathering Supr. & 305 $\pm$ 60 & \textbf{1427} $\pm$ 645 & $p<0.0001$ \\
    Deadly Corridor & 0.00 $\pm$ 0.0 & \textbf{40} $\pm$ 49 & $p<0.0001$ \\
    My Way Home & 96.69  $\pm$ 0.12 & \textbf{97.89} $\pm$ 0.01 & $p<0.0001$ \\
    Deathmatch & \textbf{4611} $\pm$ 2595 & 4062 $\pm$ 2442 & $p=0.1250$ \\
    \hline
    Deathmatch Chainsaw & 1025 $\pm$ 809 & \textbf{3750} $\pm$ 3130 & $p<0.0001$ \\
    Deathmatch Chaingun & 1487 $\pm$ 1189 & \textbf{2852} $\pm$ 2038 & $p<0.0001$ \\
    Deathmatch Shotgun & 1375 $\pm$ 941 & \textbf{1832} $\pm$ 1752 & $p=0.0226$ \\
    Deathmatch Plasma & \textbf{4538} $\pm$ 1537 & 3248 $\pm$ 2701 & $p<0.0001$ \\
    Deathmatch Rocket & 616 $\pm$ 583 & \textbf{1463} $\pm$ 1449 & $p<0.0001$ \\
    \hline
    \end{tabular}
  \label{tab:evaluation}
\caption{Shown are average scores based on evaluating the best policies found for A2C and A2C+RoE 100 times each. The best results are shown in bold. The five last rows show how the policies that were trained on the original \emph{Deathmatch} scenario generalize to five variations where only one weapon type is available. Standard deviations are shown for each experiment and two-tailed p-values from unpaired t-tests. }
\label{tab:evaluation}
\end{table}

\begin{center}
\begin{figure*}[!htb]
  \hspace*{-0.0in}
  \includegraphics[width=1.0\textwidth]{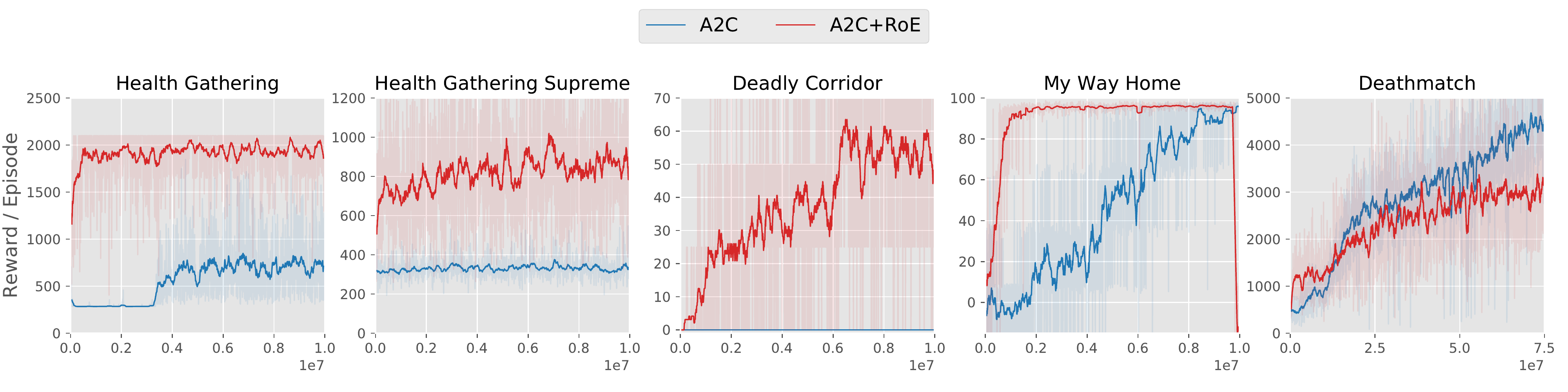} 
  \caption{The reward per episode of A2C and A2C+RoE during training in five VizDoom scenarios (smoothed). A2C is trained from the environment's extrinsic reward while A2C+RoE uses our proposed method without access to the reward. The drop in performance seen in the My Way Home scenario is discussed in-depth in Section~\ref{sec:learned-policies}.} 
  \label{fig:rewards}
\end{figure*}
\end{center}

\begin{center}
\begin{figure*}[!htb]
  \hspace*{-0.0in}
  \includegraphics[width=1.0\textwidth]{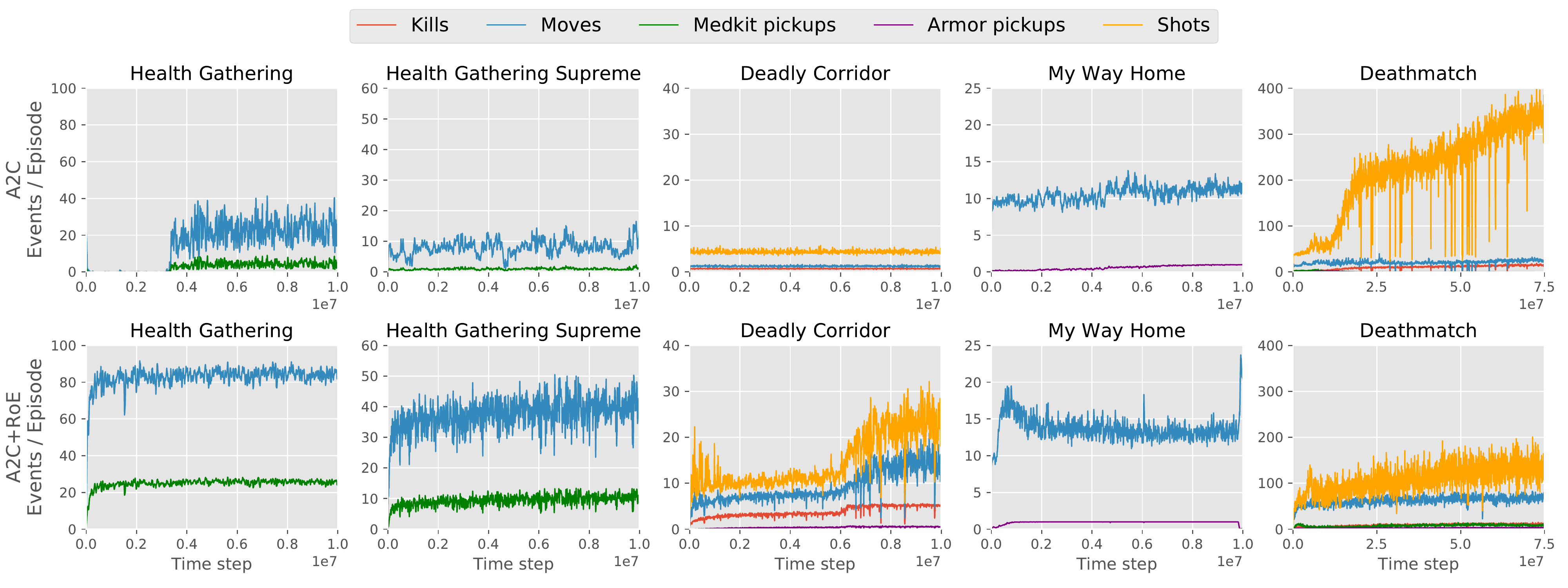} 
  \caption{Episodic mean occurrence during training for a subset of the event types in the five VizDoom scenarios. Notice the last spike in the \emph{My Way Home} scenario with A2C+RoE, in which the policy ignores the final goal (armor pickup) to prioritize continuous movement around the maze.} 
  \label{fig:events}
\end{figure*}
\end{center}

\subsection{Learned Policies}
\label{sec:learned-policies}
The A2C baseline did not learn a good policy in \emph{Health Gathering Supreme} and \emph{Deadly Corridor}, and only improved slightly in \emph{Health Gathering} (Figure~\ref{fig:rewards}). A2C learned a weak policy in three out of  five scenarios, which demonstrates that they are indeed difficult to master guided by the extrinsic rewards alone. In \emph{My Way Home}, A2C does learn a strong behavior that consistently locates and picks up the armor but only after 8--9 million training steps. In \emph{Deathmatch}, A2C learned a very high-performing behavior that directly walks to the plasma gun (the most powerful weapon in this scenario) and shoots from cover toward the center of the map. The behavior is simple but effective until it runs out of ammunition, after which it attempts to find more ammunition and sometimes fails.

Our approach A2C+RoE learns effective behaviors in all five scenarios. The learned behavior in \emph{Deathmatch} does not exclusively use the powerful plasma gun, which results in a slightly but not significantly worse performance than A2C  ($p=0.125$ using two-tailed t-test). The policy is still effective with over 10 kills per episode. These kills are spread across all weapons that are available, resulting in a behavior that is more varied (and interesting to watch).   
As we will show in Section~\ref{sec:adapt}, the versatile behavior learned by A2C+RoE allows it to adapt to critical changes in \emph{Deathmatch} in contrast to policies trained through A2C. 

The episodic mean occurrence of events (Figure~\ref{fig:events}) allows us to analyze how the policies change over time. In \emph{Health Gathering} and \emph{Health Gathering Supreme}, A2C+RoE quickly learns to move $\sim$80 and $\sim$30 units per episode, respectively. This behavior might  explain why the agent also quickly learns to pick up medkits. A2C, on the other hand, learns the relationship between movement, medkits, and survival at a much slower pace, at least in the \emph{Health Gathering} scenario. In \emph{Deadly Corridor} A2C+RoE discovers an interesting behavior. After the agent learns to kill all six enemies (the red line) and to pick up armor (purple line), it still manages to increase the movement and the shooting events; the agent learned to walk back to its initial position while shooting and then afterwards to return to pick up the armor. This result is not unexpected as the agent is intrinsically motivated to experience as many events as possible during an episode. 

In \emph{My Way Home}, after the A2C+RoE policy has learned to routinely pick up the armor, it shifts into a different behavior toward the end of training. The agent learned to avoid the armor to instead continuously move around in the maze. We suspect that the policy would shift back to the previous behavior if training was continued, as the movement reward is now decreasing and the armor reward is increasing. Since our rarity measure is temporal, loops between these two behaviors could emerge as well. As  policies with the highest extrinsic reward are saved during training, these sudden changes do not affect the final policy. In fact, one might argue that this is a useful feature of RoE: a network that has converged to some optimum can escape it to find other interesting behaviors.

\begin{center}
\begin{figure*}[!htb]
  \hspace*{-0.0in}
  \includegraphics[width=1.0\textwidth]{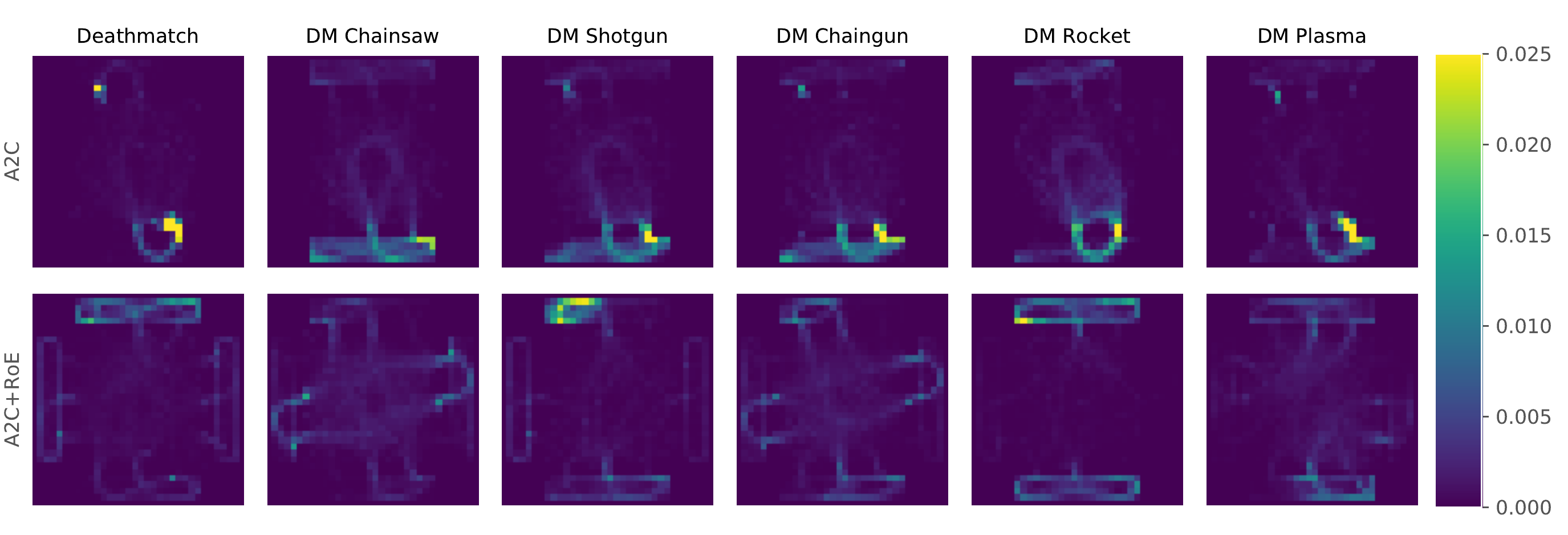} 
   \caption{Heat maps showing the proportional time spent at each location on the map in the \emph{Deathmatch} scenario and its five variations. The values are based on evaluating the two trained policies 100 times each and clipped at 0.025. The heat maps show that the A2C-policy prefers to stay near the plasma gun, even in the map variations where it is not present, while the A2C+RoE-policy has learned distinct behaviors for each weapon type. The results in Table~\ref{tab:evaluation} shows that the A2C+RoE-policy is able to reach high scores in these variations event though it was never trained on them.} 
  \label{fig:heat}
  \vspace{-1em}
\end{figure*}
\end{center}

\vspace{-0.1in}
\subsection{Ability to Adapt}
\label{sec:adapt}
A2C+RoE motivates the agent intrinsically to learn a balanced policy that strives to experience a good mix of events. Reinforcement learning algorithms that exclude pre-training or proper reward shaping, including our A2C baseline, can easily converge into local optima with very \emph{narrow} behaviors. In this context, \emph{narrow} refers to behaviors that  act in a very particular way, only utilizing a small subset of the features in the environment. This handicap prevents the learned policies from adapting to critical changes in the environment as they only know one way of behaving. 

To test for such adaptivity, the learned policies are evaluated on five \emph{Deathmatch} variations in which critical weapons and ammunition packs have been removed. Note that the policies were not directly trained on these variations. The results in Table~\ref{tab:evaluation} show that A2C+RoE learned a policy that significantly outperforms A2C ($p<0.0001$ using two-tailed t-test) in four out of five \emph{Deathmatch} variations. A2C+RoE learned a policy that is more versatile, capable of using all the weapons in the map, which is the reason it can easily adapt. Figure~\ref{fig:heat} shows heat maps (i.e.\ the proportional time spent at each map location) during the evaluations of the two policies on \emph{Deathmatch} and its variations. The A2C+RoE policy expresses different strategies depending on the weapon available on the map, while the A2C policy mostly circles around the plasma gun location, regardless of it actually being there. However, if the plasma gun is present, A2C alone does execute a fairly effective strategy, shooting toward enemies in the middle of the map.

The heat maps show that the A2C policy has learned to stay at only one location on the map from which it can pick up the powerful plasma gun and thereafter shoot efficiently toward enemies in the middle of the map (see the video of the learned policies). In the \emph{Deathmatch} variations, in which the map only contains two weapons of the same type, the A2C-policy fails to adapt to use the other weapons and instead walks around the area where the plasma gun would have been located.

The A2C+RoE policy has learned to explore a larger part of the maps in a more uniform way (Figure~\ref{fig:heat},bottom). In the different \emph{Deathmatch} variations, a clear change in behavior can be observed when only a certain type of weapon is available. For example, in the \emph{DM Rocket} scenario, the agent lures enemies into the map's top and bottom room while efficiently using the rocket's splash damage.

\section{Discussion}
While the presented approach worked well in VizDoom it will be important to test its generality in other domains in the future. RoE is designed to work well in challenging environments that have a plethora of known events and sparse and/or delayed rewards. Video games are thus a very suitable domain and we plan to test RoE in Montezuma's Revenge and StarCraft in future work. For domains in which reward shaping is not necessary, i.e.\ the extrinsic reward smoothly leads to an optimal behavior, our approach might add less value. We imagine that RoE  should also work well in domains with deceptive reward structures, just as novelty search outperformed traditional evolutionary algorithms in mazes with dead ends \cite{lehman2011abandoning} or deceptive meta-learning tasks  \cite{risi2010evolving}. Novelty search and RoE have the ability to learn interesting behaviors without the need for a goal. In the future, our approach could also be extended to reward the agent for both the rarity of events as well as the environment's original objective, inspired by \emph{quality diversity methods} \cite{pugh2016quality} that use a combination of diversity and objective-based search \cite{cuccu2011novelty, mouret2015illuminating}. 

The specification of adequate events is intimately tied to the success of our approach; events that lead to direct negative performance should be avoided. For example, if the extrinsic reward is negative when the agent wastes ammunition, it should not be intrinsically rewarded for shooting event. A benefit of the presented method is that events that contribute to the occurrence of other events (e.g.\ such as movement leads to medkit pickups),  can lead to a system that performs automated curriculum learning. However, it is not guaranteed that this effect will occur, and it might require a bit of trial and error during the specification of events.  
Some events can also be contradicting, such as killing with the chainsaw and killing with the plasma gun, as the agent cannot do both at the same time. Our approach is designed to learn a policy that can balance their occurrences which results in a more versatile behavior. Important future work will test how RoE scales to hundreds or even thousands of events. A promising testbed for such experimentation is StarCraft, for which events can easily be defined as the production of each unit and building type, as well as killing different opposing unit types. We believe that reinforcement learning methods that are guided by intrinsic motivation are key to solving these challenging environments. 

The A2C baseline reached the best performance in the original \emph{Deathmatch}. However, it can be argued whether it learned to actually play Doom, or  just learned to follow a fixed sequence of actions that lead to the same behavior every time. While it can be useful to find a niche behavior with high performance, learning a rich and versatile behavior has particular relevance for video games. Here, behaviors that explore the game's features could potentially help for automatic game testing and also lead to more human-like behaviors  for NPCs.


Regarding our implementation of the RoE approach, future work will also explore other variations in determining the episodic mean occurrence of events, such as discounting the mean occurrences over time. With this modification, event occurrences older than $N$ episodes (the event buffer only holds $N$ event occurrences) would still effect the intrinsic reward. 


It is important to note that since we save the best model based on the mean extrinsic reward across all worker threads, increasing the number of threads should make the evaluation less noisy by reducing the chances of accidentally overriding the best model with a worse performing one. This hypothesis still needs to be confirmed, but the number of threads was already increased from 4 to 16 in the longer \emph{Deathmatch} scenario to speed up learning. 


\section{Conclusion}
We introduced \emph{Rarity of Events} (RoE), a simple reinforcement learning approach that determines reward based on the temporal rarity of pre-defined events. This approach was able to reach high-performing scores in five challenging VizDoom scenarios with sparse and/or delayed rewards. 
Compared to a traditional A2C baseline, the results are significantly better in four of the five scenarios. Importantly, the presented approach is able to not only receive a high final reward, but also discovers versatile behavior that can adapt to critical changes in the environment, which is challenging for the baseline A2C approach. In our experiments, the extrinsically motivated baseline either fails in these environments or learns a 
behavior that is unable to adapt to changes in the environments it has been trained on. In the future, the presented RoE approach could allow more complex scenarios to be solved, for which it is infeasible to learn from extrinsic rewards without manual reward shaping and curriculum learning.

\section{Acknowledgements}
We thank OpenAI for publishing accessible implementations of A2C and GitHub user p-kar for the integration of A2C to VizDoom\footnote{https://github.com/p-kar/a2c-acktr-vizdoom}. We would also like to show our gratitude to the members of the Game Innovation Lab at New York University Tandon School of Engineering for their feedback and inspiring ideas. This work was financially supported by the Elite Research travel grant from The Danish Ministry for Higher Education and Science.

\bibliographystyle{abbrvnat}

{\small 
\bibliography{references.bib}
}

\end{document}